\documentclass{article}

\PassOptionsToPackage{numbers, compress}{natbib}
\usepackage[preprint]{neurips_2025}

\usepackage[utf8]{inputenc} 
\usepackage[T1]{fontenc}    
\usepackage{hyperref}       
\usepackage{url}            
\usepackage{booktabs}       
\usepackage{amsfonts}       
\usepackage{nicefrac}       
\usepackage{microtype}      
\usepackage{xcolor}         
\usepackage{graphicx}
\usepackage{amsmath}
\usepackage{cleveref}
\usepackage{wrapfig}

\title{Benchmarking Unsupervised Strategies for Anomaly Detection in Multivariate Time Series}

\author{%
  Laura Boggia\thanks{Corresponding author: \href{mailto:laura.sara.boggia@ibm.com}{laura.sara.boggia@ibm.com}\\Code available at: \url{https://github.com/laurab222/TSAD}.
  This paper is currently under review for the VLDB 2026 conference.}\\
  IBM Research Paris-Saclay\\
  LPNHE, Sorbonne Université, Université Paris Cité, CNRS/IN2P3\\
  Paris, France\\
  \And
  Rafael Teixeira de Lima\\
  IBM Research Paris-Saclay\\
  Paris, France\\
  \AND
  Bogdan Malaescu\\
  LPNHE, Sorbonne Université, Université Paris Cité, CNRS/IN2P3\\
  Paris, France\\
}

\begin{document}

\maketitle

\begin{abstract}
Anomaly detection in multivariate time series is an important problem across various fields such as healthcare, financial services, manufacturing or physics detector monitoring. 
Accurately identifying when unexpected errors or faults occur is essential, yet challenging, due to the unknown nature of anomalies and the complex interdependencies between time series dimensions.

In this paper, we investigate transformer-based approaches for time series anomaly detection, focusing on the recently proposed iTransformer architecture.
Our contributions are fourfold: (i) we explore the application of the iTransformer to time series anomaly detection, and analyse the influence of key parameters such as window size, step size, and model dimensions on performance; (ii) we examine methods for extracting anomaly labels from multidimensional anomaly scores and discuss appropriate evaluation metrics for such labels; (iii) we study the impact of anomalous data present during training and assess the effectiveness of alternative loss functions in mitigating their influence; and (iv) we present a comprehensive comparison of several transformer-based models across a diverse set of datasets for time series anomaly detection. 
\end{abstract}

\section{Introduction}
Anomaly detection, i.e. the identification of specific data instances that do not conform to the vast majority of it, is an important problem across a wide range of domains. 
This task is challenging due to the unknown nature of these instances, with respect to the majority, and their aforementioned rarity.
However, interest in anomaly detection continues to grow due to its valuable applications for real-world problems such as fraud detection, intrusion detection, medical diagnosis, risk prediction, and many more~\cite{aggarwal_outlier_2017}. 

With the evolution of the Internet of Things and the increase in data taking in general, many relevant data from various applications come in the shape of time series, i.e. sequential data instances indexed over time~\cite{zamanzadeh_darban_deep_2024}. 
Time Series Anomaly Detection (TSAD) is a critical task in diverse fields such as healthcare, financial services, cybersecurity, transportation, and monitoring systems, where it ensures the correct operation of manufacturing lines, water treatment facilities or physics detectors, to name a few.
Accurate identification of the moment when unexpected errors or defects occur is essential for these applications. 
One typically distinguishes between \textit{univariate} (one-dimensional) and \textit{multivariate} (multidimensional) time series.
In this case, the correlations between different dimensions must be taken into account, which increases the complexity of the problem.

Anomaly detection is typically approached in an \textit{unsupervised} manner, as the nature of anomalous instances is not defined beyond their deviation from the majority of the data.
However, in certain well-studied settings, where anomalies are properly characterised, the data can be labelled and \textit{supervised} learning methods become applicable.
In reality, due to the rare occurrence and ill-defined nature of anomalies, obtaining reliable labels is often difficult and costly.
Consequently, supervised learning is unsuitable for many real-world applications, and one relies on \textit{unsupervised} learning.
Furthermore, one distinguishes between \textit{unsupervised} and \textit{self-supervised} methods.
Whilst both work without anomaly labels, \textit{self-supervised} models create their own targets from the input data and train with those.
However, \textit{unsupervised} and even \textit{self-supervised} techniques often perform worse than the supervised approach ~\cite{zamanzadeh_darban_deep_2024, han_adbench_2022} when labels are available.

In this work, we focus on unsupervised techniques for TSAD, due to the previously mentioned challenges associated with supervised approaches, and discuss the end-to-end setup of a TSAD algorithm.
We make sure to include a wide choice of data from various applications to avoid biasing our studies by some of the limitations of current benchmark datasets.

First, we give a brief overview over related work on TSAD (\Cref{sect:rel_work}) and then present the datasets (\Cref{sect:datasets}), algorithms (\Cref{sect:algo}) and experimental setup (\Cref{sect:setup}) used for our studies.
Finally, \Cref{sect:res} presents and discusses our results. 
Our main contributions are:
\begin{itemize}
    \item exploring the use of the \texttt{iTransformer} ~\cite{liu_itransformer_2024} architecture for anomaly detection, and  studying the influence of window, step and internal model size on the anomaly detection performance;
    \item analysing how to extract anomaly labels from a multidimensional anomaly score provided by an anomaly detection model, and what metric to use to evaluate the extracted labels;
    \item further investigating the impact of anomalies present in training data and how to mitigate this using alternative loss functions;
    \item comparing different transformer-based models for TSAD on a wide range of datasets.
\end{itemize}

\section{Related Work}
\label{sect:rel_work}
\subsection{Anomaly Detection Algorithms}
TSAD has become an increasingly active area of research. 
With the growing availability of data from diverse sensors and monitoring systems, numerous new applications have emerged, giving rise to a wide range of novel research directions.
Given the vast range of applications and corresponding research in these fields, we will not cover them entirely but refer the reader to more exhaustive surveys and reviews~\cite{zamanzadeh_darban_deep_2024, gupta_outlier_2014, chalapathy_deep_2019, blazquez-garcia_review_2021, schmidl_anomaly_2022}.
Furthermore, for the specific topic of univariate TSAD, more information can be found in~\cite{braei_anomaly_2020, paparrizos_tsb-uad_2022}.

By definition, anomalies correspond to unexpected and rare instances in data, often corresponding to patterns that have never occurred before, making it impossible to teach a model about them in advance.
This inherent unpredictability renders supervised approaches unsuitable for TSAD, leading most TSAD methods to rely on unsupervised strategies instead.

Classical, unsupervised approaches entail methods such as isolation forests (IFs)~\cite{liu_isolation_2008}, one-class support vector machines (OCSVMs)~\cite{scholkopf_support_1999},
clustering algorithms, k-nearest neighbour (kNN)~\cite{cover_nearest_1967} methods or algorithms like MERLIN ~\cite{nakamura_merlin_2020} to name a few ~\cite{lai_revisiting_2021, chandola_anomaly_2009}.
However, in recent years, many of them have been outperformed by deep learning approaches, though some remain relevant. 
Therefore, the focus in TSAD currently mostly lies on deep learning, as numerous prior works have demonstrated its superior performance ~\cite{zamanzadeh_darban_deep_2024, audibert_usad_2020, tuli_tranad_2022, deng_graph_2021, zhao_multivariate_2020}.

A complete review of deep learning approaches for TSAD can be found in ~\cite{chalapathy_deep_2019}.
Unsupervised, deep learning methods for TSAD entail models such as generative adversarial networks (GANs) ~\cite{goodfellow_generative_2014}, autoencoders (AEs) ~\cite{lecun_modeles_1987, bourlard_auto-association_1988, zemel_developing_1993}, variational autoencoders (VAEs) ~\cite{kingma_auto-encoding_2013}, graph neural networks (GNNs) ~\cite{scarselli_graph_2009} and recurrent neural networks (RNNs)~\cite{mcculloch_logical_1943, rumelhart_learning_1986} including long short-term memory (LSTM) models ~\cite{hochreiter_long_1997}.
A popular example for TSAD is \texttt{DAGMM}~\cite{zong_deep_2018} that jointly trains a Gaussian mixture model with an autoencoder.
The autoencoder helps with data compression, allowing the Gaussian mixture model to better capture multivariate correlations, whilst the combined training assures that this is done effectively.
\texttt{OmniAnomaly}~\cite{su_robust_2019} is an example of a model employing stochastic RNNs for robust anomaly detection on multivariate time series.
Next, \texttt{USAD}~\cite{audibert_usad_2020} is based on adversarially trained autoencoders, facilitating fast training and increased robustness.
Finally, an LSTM-based autoencoder~\cite{hsieh_unsupervised_2019} combines the advantages of autoencoders with the strengths of LSTMs at capturing temporal dependencies.

Recently, transformer models have received increased attention due to their ability to model long-term dependencies through self-attention mechanisms, especially when combined with positional encoding. 
They also enable parallel computation and help mitigate gradient-related issues that can still pose challenges for models like LSTMs on longer sequences ~\cite{zamanzadeh_darban_deep_2024, wen_transformers_2023, ahmed_transformers_2023}.
Transformer-based approaches to TSAD entail models such as \texttt{TranAD}~\cite{tuli_tranad_2022}, \texttt{MT-RVA}~\cite{wang_variational_2022}, \texttt{TransAnomaly}~\cite{zhang_unsupervised_2021}, \texttt{GTA}~\cite{chen_learning_2022}, \texttt{Anomaly Transformer}~\cite{xu_anomaly_2022}, \texttt{VTT}~\cite{kang_transformer-based_2024},
which explore the combination of transformers with adversarial training, VAEs, GNNs, association discrepancy or alternative embeddings, respectively.

\subsection{Benchmarks}
\label{sect:benchmarks}
Although the importance of benchmark databases is well established in the machine learning community, there is a severe lack of reliable and challenging anomaly detection data for time series.
An overview of currently available datasets is presented by \citet{lai_revisiting_2021}, but all except four of the presented datasets are synthetic.
Other public datasets that have been proposed to evaluate anomaly detection in time series are the \texttt{Yahoo} ~\cite{n_laptev_s_amizadeh_and_y_billawala_s5_2015}, \texttt{Numenta} ~\cite{ahmad_unsupervised_2017}, \texttt{NASA} (\texttt{MSL} and \texttt{SMAP}) ~\cite{hundman_detecting_2018} and \texttt{SMD} ~\cite{su_robust_2019}.
However, concerns have been raised about the validity and usefulness of these datasets for benchmarking.
~\citet{wu_current_2021} argue that most of these commonly used time series have such obvious anomalies that they can be identified by simple rules or one-line algorithms, and therefore, are not suited to benchmark a model's anomaly detection performance.
Further, they propose the \texttt{UCR} time series anomaly archive ~\cite{wu_current_2021}. 
Whilst containing synthetic but realistic anomalies, this data collection has the drawback that it is currently limited to univariate time series.
Similarly, ~\citet{nakamura_merlin_2020} assert that current benchmarks are insufficient and do not allow for a complete comparison of emerging algorithms.
Finally, ~\citet{han_adbench_2022} focuses on tabular data for anomaly detection but includes some recommendations for TSAD.

\section{Datasets}
\label{sect:datasets}

\subsection{Public Data for TSAD}
Despite the previously discussed issues with current benchmarks and ongoing efforts for more challenging datasets (see \Cref{sect:benchmarks}), many of the benchmark sets mentioned before are still widely used in TSAD.
To facilitate comparison with older models, some are still included in this study.
Namely, we use the \texttt{SMD}~\cite{su_robust_2019} dataset containing server machine data, and the two NASA datasets, \texttt{MSL} and \texttt{SMAP}~\cite{hundman_detecting_2018}, which are both based on sensor data collected on Mars.
We add the new \texttt{UCR} which was proposed by ~\citet{wu_current_2021}, to mitigate some of the issues of the datasets above.
Next, we also include the frequently used \texttt{SWaT}~\cite{mathur_swat_2016} and \texttt{WADI}~\cite{ahmed_wadi_2017} datasets, both incorporating data from monitoring water treatment plants, and \texttt{SWaT\_1D}~\cite{julienau_anomaly_2020}, a univariate time series derived from \texttt{SWaT}.
Additionally,~\citet{lai_revisiting_2021} mention the \texttt{Credit Card} ~\cite{pozzolo_calibrating_2015} and the \texttt{GECCO} ~\cite{bartz-beielstein_internet_2018} datasets as examples of real-world benchmarks.
These datasets are selected for their diverse anomaly types: The \texttt{Credit Card} set features point anomalies corresponding to fraudulent credit card transactions, while the \texttt{GECCO} set shows collective anomalous patterns signalling an unsafe water quality. 
Finally, we introduce the \texttt{IEEECIS}~\cite{howard_ieee-cis_2019, tapia_fraud_2022}, a dataset consisting of online transactions with a challenging combination of point and collective anomalies.
With this selection, we aim to have a broadly varied spectrum of application fields, anomaly types and data characteristics.
These datasets are described in \Cref{tab:public_data}, and additional information can be found in \Cref{appendix:data_more}.

\begin{table}[ht]
    \caption{Public datasets used for TSAD. The values in brackets next to the anomaly rate indicates the number of anomalous sequences in the test data. The average, maximum and minimum anomaly lengths are measured on the labelled test data.\\}
    \label{tab:public_data}
    \centering
    \resizebox{\textwidth}{!}{%
    \begin{tabular}{@{}l|l|ccccccc@{}}
    \toprule
    Dataset & Description & Dimensions & Train size & Test size & \begin{tabular}[c]{@{}c@{}}Anomaly rate \\ (\# anomalous seq.)\end{tabular} & \begin{tabular}[c]{@{}c@{}}Avg. anom. \\ length\end{tabular} & \begin{tabular}[c]{@{}c@{}}Max. anom. \\ length\end{tabular} & \begin{tabular}[c]{@{}c@{}}Min. anom. \\ length\end{tabular} \\ \midrule
    Credit card & Credit card transactions & 29 & 200k & 85k & 0.13\% (107) & 1 & 2 & 1 \\
    GECCO & Water quality & 9 & 70k & 70k & 1.05\% (22) & 33 & 157 & 23 \\
    IEEECIS & Online payments & 178 (use 30) & 17k & 1k & 6.67\% (12) & 5 & 21 & 1 \\
    MSL & Sensors of Mars rover & 55 & 58k & 74k & 10.48\% (36) & 215 & 1140 & 10 \\
    SMAP & Soil moisture on Mars & 25 & 138k & 436k & 12.82\% (68) & 821 & 4217 & 30 \\
    SMD & Server machine & 38 & 105k & 105k & 5.14\% (35) & 154 & 837 & 2 \\
    SWaT & Water treatment & 51 & 495k & 450k & 12.13\% (35) & 1560 & 35900 & 101 \\
    SWaT\_1D & Water quality & 1 & 3k & 5k & 12.88\% (83) & 8 & 211 & 1 \\
    UCR & Healthcare & 1 & 1.6k & 5.9k & 1.88\% (4) & 59 & 111 & 10 \\
    WADI & Water treatment & 127 (use 30) & 785k & 173k & 5.77\% (14) & 713 & 1741 & 88\\
    \bottomrule
    \end{tabular}%
    }
\end{table}

\subsection{Types of Anomalies}
\begin{wrapfigure}{l}{0.55\linewidth}
    \centering
    \includegraphics[width=\linewidth]{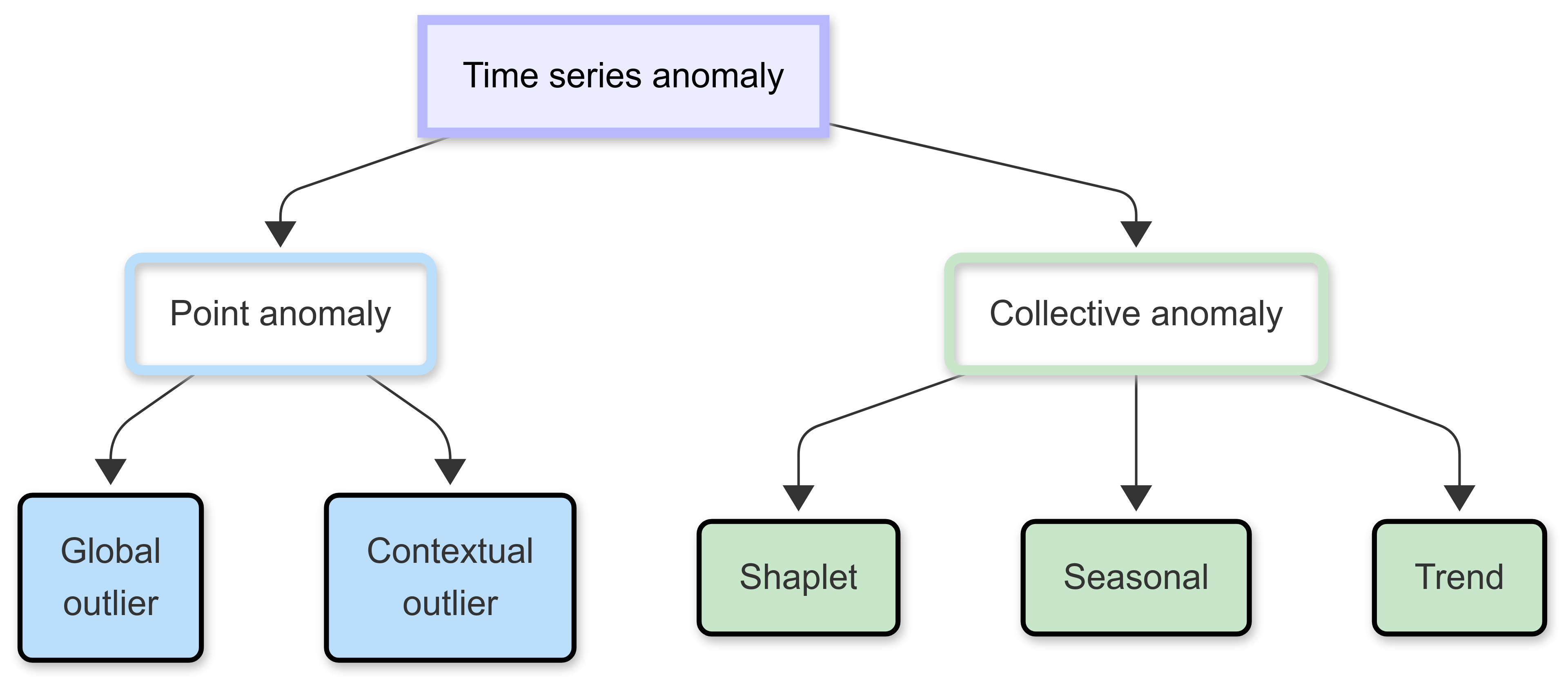}
    \caption{Taxonomy for anomalies in time series.}
    \label{fig:schema_TSanomalies}
\end{wrapfigure}

In TSAD, one typically distinguishes between \textit{point} and \textit{collective} anomalies.
While the taxonomy of time series anomalies may differ slightly depending on the exact application, the fundamental definitions remain consistent.
A visual characterisation of them is given in \Cref{fig:schema_TSanomalies} and further discussion can be found in ~\cite{zamanzadeh_darban_deep_2024, gupta_outlier_2014, blazquez-garcia_review_2021,paparrizos_tsb-uad_2022, lai_revisiting_2021, pang_deep_2022}. 

First, \textit{point} anomalies describe anomalies that are highly-localised and correspond to one anomalous time stamp.
We further distinguish anomalies corresponding to an extreme point of the time series, which are called \textit{global} outliers, from \textit{contextual} anomalies, that describe time stamps that only appear anomalous given the context of their neighbouring time stamps.\\
Second, \textit{collective} anomalies are also called subsequence or group anomalies, as they encompass a longer, anomalous sequence of a time series, containing multiple time stamps that are anomalous.
Often, each of the individual time stamps of such a collective anomaly appears normal, but together they create an anomalous pattern.
Such anomalous sequences can be characterised by having either a different pattern shape, trend or seasonality than the previous part of the time series.\\
Since contextual and collective anomalies both depend on temporal context, they are occasionally treated as a single category, as done in ~\cite{hundman_detecting_2018}.

\Cref{tab:public_data} describes the anomaly characteristics such as the type, rate and average, maximum and minimum length of anomalies for each dataset.
The given numbers only concern the testing set and not the training set, as in most cases the training data are either provided without labels or do not contain anomalies by construction.

\section{Existing Algorithms}
\label{sect:algo}
We first introduce a deterministic baseline approach and a non-transformer model for comparison, followed by transformer-based model candidates for TSAD.

\subsection{Baseline Approach}
\label{sect:baseline}
As detailed previously, we do not use supervised models, focusing on strategies that do not rely on labelled data for training.
To provide a simple baseline algorithm that satisfies that requirement, we directly take the absolute value of the input time series as an anomaly score. 
This approach is described in ~\cite{kim_towards_2022} and recommended for better comparison against other established anomaly detection models.
Different anomaly label extraction strategies (explained in \Cref{anomaly_label_methods}) can be applied to those anomaly scores.
The baseline, already achieving a high anomaly detection score, implies that the information in the unprocessed time series is already sufficient to find anomalies, and more complex learning models are unnecessary.

\subsection{Deep Learning Approaches}
Focusing on more complex models, we categorise deep learning models based on their underlying assumptions.
The vast majority of anomaly detection models are \textit{reconstruction-based}.
In this case, the input data is fully reconstructed by the model, and the difference between input and reconstruction is used to identify anomalies (anomaly score). 
This reconstruction error is typically quantified by the mean-square error (MSE).
In the context of anomaly detection, the fundamental assumption is that normal data, since abundant and therefore well-learned, are easy to reconstruct, whereas anomalous instances entail higher reconstruction errors.

Another popular assumption for time series anomaly detection is that anomalies are harder to predict than normal instances, i.e. a well-trained forecasting model would have larger prediction errors for anomalies than normal data, and its prediction error can hence be used as an anomaly score. 
Furthermore, \textit{forecasting-based} anomaly detection might be sensitive to other types of anomalies than reconstruction-based.
For instance, it might be harder for a forecasting-based model to predict contextual anomalous data in a rapidly changing time series than for a reconstruction-based model ~\cite{zamanzadeh_darban_deep_2024}, leading to larger forecasting errors.

\subsubsection{Deep Learning Baseline Approach} \hfill\\
\label{sect:usad} 
As a deep learning baseline approach, we choose the \texttt{USAD}~\cite{audibert_usad_2020} model.
It tackles multivariate TSAD by combining an autoencoder with adversarial training to improve the robustness and efficiency of the training.
It is a reconstruction-based anomaly detection model, but instead 
of solely focusing on the reconstruction error (measured by the MSE), \texttt{USAD} splits the training into two phases.

In the first phase, the compression and reconstruction are learned with the MSE.
The \texttt{USAD} uses a single encoder and two decoder models, combining them into two autoencoder models $AE_1, AE_2$ with the same encoder.
Both autoencoders are trained identically during this phase.
In the second phase, the two autoencoders are depicted against each other, where data reconstructed by $AE_1$ is used to try to fool $AE_2$, which tries to distinguish real data from reconstructed data.
During testing, the anomaly score is defined as 
\begin{equation}
    l(y) = \alpha | y - AE_1(y)| + \beta |y - AE_2(AE_1(y))|
\end{equation} where $\alpha + \beta =1$.

\subsubsection{Transformer-based Anomaly Detection} \hfill\\
Transformers are known for dealing exceptionally well with sequential data such as language.
Based on this observation, current research tries to determine how far this applies to other types of sequential data, including time series. 
Discussions concerning the usefulness of transformers for time series forecasting are still ongoing ~\cite{ferrari_dacrema_are_2019, elsayed_we_2021, zeng_are_2022, das_long-term_2024}, as ~\citet{zeng_are_2022} produced better results with linear models than with transformer-based models.
However, unlike traditional sequential models such as RNNs and LSTMs, transformers are not susceptible to vanishing gradients or inefficiencies in computation. 
Furthermore, their self-attention mechanism enables them to capture complex, multidimensional dependencies in parallel, making them highly effective for modelling long-range relationships in sequential data.
For these reasons, various transformer-based models for time series have emerged in the last few years, and we will focus on two recent and interesting models while also including a \texttt{vanilla Transformer} model for comparison. 

First, ~\citet{tuli_tranad_2022}, the authors of the \textbf{TranAD} model, tackle multivariate TSAD by introducing adversarial training and self-conditioning.
The adversarial training procedure helps amplify the reconstruction error for subtle anomalies, whereas the self-conditioning increases the focus on anomalous subsequences.

Next, the inverted Transformer (\textbf{iTransformer}) model presented by ~\citet{liu_itransformer_2024} was initially introduced for long-term time series forecasting.
The authors argue that it's not the transformer itself that is not apt for forecasting multivariate time series, but the standard approach of data encoding.
Inverting the data and thereby switching the roles of time and variates, forces the model to apply its attention mechanism across the variates instead of the time dimension, which allows to learn multivariate connections while conserving the temporal dependencies.
Further, adding a feed-forward layer after the attention layer encourages the model to learn a representation of the time series.
It has been shown that this inverted embedding improves the results for long-term time series forecasting remarkably.\\
A naturally emerging research problem is to test this approach on anomaly detection.
As a first step, it is straightforward to use the \texttt{iTransformer} model for forecasting-based anomaly detection.
Furthermore, we test whether the inverted data encoding data also proves to be useful for reconstruction and reconstruction-based anomaly detection.

Finally, the \textbf{(vanilla) transformer} refers to the standard transformer architecture introduced by~\citet{vaswani_attention_2017}.
Just using the encoder of said model and plugging its outputs into a feed-forward network, can provide a simple reconstruction-based anomaly detection model.
To not lose the temporal aspect of the data, positional encoding is used when embedding the data. 
Multiple transformers have already been studied for AD~\cite{zamanzadeh_darban_deep_2024, wen_transformers_2023, ahmed_transformers_2023}. \\
The \texttt{vanilla Transformer} model is introduced to study the effect of the inverted embedding used for the \texttt{iTransformer}, against the standard embedding of the \texttt{vanilla Transformer}.
It has been shown in the \texttt{iTransformer} paper~\cite{liu_itransformer_2024} that for long-term time series forecasting, the inverted embedding shows a clear advantage in terms of prediction performance and flexibility over the standard approach.
However, it is not evident whether this advantage holds in the case of anomaly detection, and we therefore compare the \texttt{iTransformer} to the \texttt{vanilla Transformer} to assess the impact of the inverted embedding, within the context of reconstruction-based TSAD.

\subsection{Anomaly Label Extraction Strategies} 
\label{anomaly_label_methods}
The following paragraphs detail how to obtain actual binary anomaly labels starting from an anomaly score.
In the first step, we focus on univariate anomaly scores before extending our approach to multivariate scores.

\subsubsection{Anomaly Labels for Univariate Time Series}\hfill\\
A possible, static labelling approach is studying the anomaly scores on a validation set with corresponding anomaly labels and determining the threshold that yields the best value of a chosen metric on this set. 
The same threshold is then used for the test set. 
The downside of this approach is the need for a labelled holdout set, in a setting where labelled data are already scarce ~\cite{kim_towards_2022}. \\
If one has an approximate idea of the fraction $f$ of anomalies present in the data, it can be used to declare the $f$ percentile of the highest anomaly scores as anomalous to derive a threshold from there. \\
Finally, the Peak-over-Threshold (POT)~\cite{siffer_anomaly_2017}  method allows to derive a threshold based on Extreme Value Theory (EVT)~\cite{beirlant_statistics_2005} and is commonly used to identify extreme points from data without any strong assumptions on the underlying data distribution.

\subsubsection{Anomaly Labels for Multivariate Time Series}\hfill\\
The challenge with multivariate time series is effectively managing the complexity introduced by the multi-dimensional nature of the data.
Typically, models provide anomaly scores separately for every variate of the time series, i.e. a time series with $N$ variates yields an $N$-dimensional anomaly score.
However, for most applications, a single label per time stamp is more practical, regardless of which variate is anomalous.
This raises two key questions: how should these scores be combined, and how does the choice of combination method affect anomaly detection performance?
We distinguish between \textit{global} and \textit{local} combination strategies and compare three such approaches. 
Note that in the case of univariate time series, all methods produce identical anomaly labels.
\begin{itemize}
    \item \textbf{global}: The anomaly scores are first averaged over all $N$ variates to get a univariate series of anomaly scores to which the previously described anomaly label extraction strategies can be applied.
    \item \textbf{local (inclusive OR)}: The anomaly label extraction strategies are applied separately to each variate of the time series. Next, to end up with only one label per time stamp, we apply an \textit{inclusive OR}-pooling function across the $N$ variates, i.e. if a time stamp is anomalous in at least one of the $N$ dimensions, the time stamp is anomalous in the final labels.
    \item \textbf{local (majority voting)}: Again, the anomaly label extraction strategies are applied separately to each variate of the time series and then we apply \textit{majority voting} across the variates, i.e. if at least $N/2$ dimensions are anomalous at a certain time stamp, the time stamp is declared anomalous in the combined labels. 
\end{itemize}

In prior work, the \textit{global} method is the most prevalent approach for dealing with multidimensional anomaly labels~\cite{tuli_tranad_2022, kang_transformer-based_2024}, though other options have been explored~\cite{hundman_detecting_2018, li_robust_2018}.
It is also possible to explore alternative pooling functions for local anomaly labels, rather than limiting oneself to inclusive OR or majority voting as done in the context of this work.
Finally, the anomaly score combination strategies presented here can be further enhanced by incorporating domain knowledge.
For instance, a practitioner may assign higher weights to individual input variates if those variates are considered more reliable. 
Additionally, point-wise estimates of the reconstruction uncertainty can be leveraged to perform a weighted combination of anomaly scores.

\section{Metrics and Experimental Setup}
\label{sect:setup}
\subsection{Sliding Windows}
\begin{figure}[ht!]
    \centering
    \includegraphics[width=\linewidth]{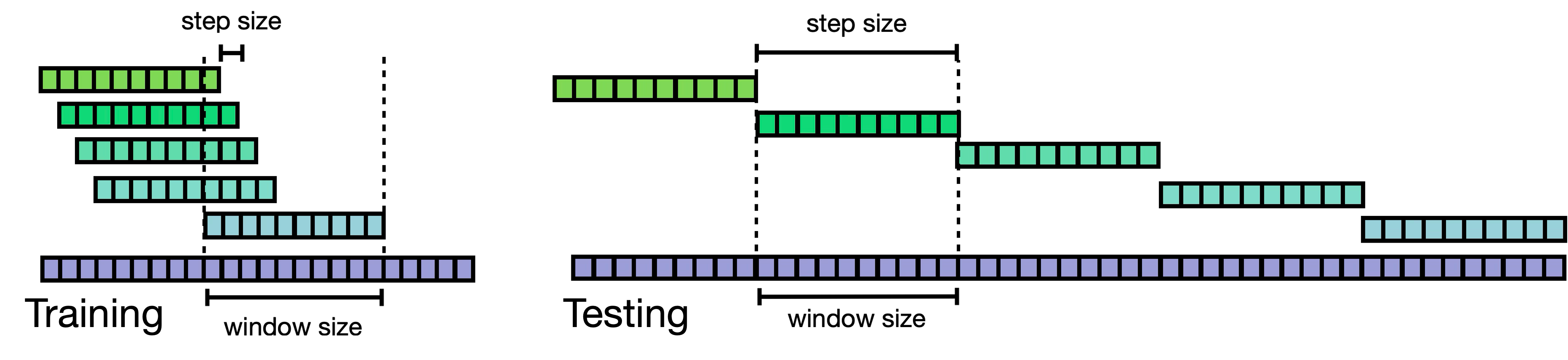} 
    \caption{Sliding window protocol for training and testing.}
    \label{fig:sliding_window}
\end{figure}

Most models that are applied to time series operate on smaller, fixed-sized segments of said series.
For this, the time series is sliced into equally sized windows of size $W$ as shown in \Cref{fig:sliding_window}. 
The overlap between consecutive windows is determined by the step size $S$.
To keep the window size consistent, the last window is padded by repeating the last value.

During training, the windows are passed to the model as inputs. 
For forecasting-based models, it is beneficial to set $S=1$, such that the model only has to predict the last time stamp per window. 
As $S$ increases, the model has to predict multiple time stamps simultaneously, which typically decreases performance due to increased difficulty.
The training loss is generally given by the mean squared error (MSE) between the predicted and actual last time stamp of the window and then averaged over all windows.
The same configuration is also used for testing, except here, the MSE is not averaged but retained as a multi-dimensional anomaly score.

In contrast, reconstruction-based models allow for more flexibility when adjusting the step size.
During training, each window is passed through the model, which compresses and then reconstructs it.
The loss is defined as the MSE between the input window and the reconstructed output. 
For each training epoch, the loss is averaged over all windows.
For testing, the windows are chosen to be non-overlapping (i.e. $S=W$) and concatenating the MSE of each window gives the desired, multi-dimensional anomaly score.

\subsection{Detection Performance Metrics}
Once the predicted labels are obtained, the next difficulty is how to evaluate and compare them against the true labels.
For this, various metrics have been discussed.

For balanced classification problems, the classifier's performance is quantified either with the accuracy or the area under the curve (AUC) of the receiver operating characteristic (ROC) curve~\cite{powers_evaluation_2020}.\\
However for very imbalanced problems such as most anomaly detection scenarios, these metrics tend to grossly overestimate a model's performance and should not be used.
Therefore, studies often employ either precision, recall or a combination, such as the F1 score.
Note that the F1 score gives the most weight to the true positives and can hence be a rather conservative estimation of a model's performance when neglecting the true negatives~\cite{kim_towards_2022}.\\
Finally, the Matthews correlation coefficient (MCC) is also often employed, as it gives equal weight to the true positives and negatives. It is defined in the following way: 
\begin{equation}
    MCC = \frac{TP \cdot TN - FP \cdot FN}{\sqrt{(TP+FP)(TP+FN)(TN+FP)(TN+FN)}},
    \label{mcc}
\end{equation}
where TP, TN, FP, and FN denote the number of true positives, true negatives, false positives and false negatives, respectively.
Since it is a correlation coefficient, the MCC ranges between -1 and 1 and isn't strictly positive as the other previously mentioned metrics~\cite{chicco_advantages_2020}.
For the studies presented in this paper, we will focus on the MCC score.

\subsection{Alternative Loss Functions}
\label{sect:loss_fct}
The mean squared error (MSE) is the most commonly used loss function for anomaly detection. 
However, we observe that for some anomalies, TSAD models are able to learn the patterns sufficiently well to achieve low reconstruction or forecasting errors. 
This suggests that MSE may not be the optimal metric for calculating anomaly scores.
Furthermore, in the case where anomalies are included in the training data, it might be beneficial to use a loss function that is less sensitive to outliers than the MSE.
Therefore, we investigate the use of alternative loss functions such as the Huber loss and the soft Dynamic Time Warping (Soft-DTW) loss ~\cite{cuturi_soft-dtw_2018}.

The Huber loss~\cite{huber_robust_1964} is a continuous combination of the MSE and the absolute value error, which makes it more robust against outliers deviating from the bulk part of the distribution by a large amount.
It is given by
\begin{equation}
    l_\delta (\hat{y}, y) = 
\left\{
    \begin{array}{ll}
        \frac{1}{2}(y - \hat{y})^2, & \text{if } |y - \hat{y}| < \delta, \\
        \delta \cdot (|y - \hat{y}| - \frac{1}{2}\delta ) , & \text{otherwise},
    \end{array}
\right.
\end{equation}
where $y$ denotes the input and $\hat{y}$ the predicted or reconstructed outputs.

Dynamic Time Warping (DTW) ~\cite{sakoe_dynamic_1971, sakoe_dynamic_1978, garreau_metric_2014} is a similarity measure specifically designed for time series.
It has been utilised in kernel or nearest neighbour methods to compare and align time series sequences effectively.
However, due to its non-differentiability, it cannot be directly optimised within a deep learning model as a loss function.
The authors of ~\cite{cuturi_soft-dtw_2018} introduced the Soft-DTW, which is differentiable, and use it as a loss function for time series forecasting.\\
Here, we test whether the Soft-DTW loss can be useful for TSAD. 
Since Soft-DTW does not provide an element-wise loss value, it can only be used during training, and we again use the MSE as the anomaly score during testing.

\subsection{Model Configurations}
Whenever possible, we follow the model configuration as detailed in their respective papers.
The baseline, introduced in \Cref{sect:baseline}, is not based on any learning model and doesn't need any specific configuration.

Since the \texttt{iTransformer} architecture has not yet been explored for anomaly detection, we take this opportunity to evaluate various configurations for both the reconstruction- and forecasting-based approaches.
We focus on the relations between the window size $W$, the step size $S$ and the internal model size $M$ and aim to find the configuration that yields the best MCC scores, averaged over five random seed repetitions, for each dataset.
For the forecasting-based approach, we reduce the number of tested configurations to prioritise efficiency and robustness of the model.
We also use this parameter search to compare the impact of the different anomaly label extraction strategies presented in \Cref{anomaly_label_methods}.

For each parameter configuration, we compute the MCC scores using one of the three anomaly labelling methods and select the configuration with the highest score.
If the highest MCC score across five repetitions, along with its standard deviation, overlaps with the scores from other configurations, the configuration with the smallest model size is preferred.
If multiple optimal configurations remain, the one with the smallest step size is selected. 
This choice often aligns with the configuration that has a smaller standard deviation, indicating greater robustness.
Further details of this study can be found in \Cref{appendix:model_config}.

\section{Experimental Results}
\label{sect:res}
\subsection{iTransformer Model Configurations} 
\label{sect:iTransformer}
We perform a parameter search for the best configuration for the \texttt{iTransformer} model, once when the model is used for reconstruction-based anomaly detection (\texttt{iTransformer-reco}) and once for forecasting-based anomaly detection (\texttt{iTransformer-fc}).
The optimal configuration choices for each dataset are presented in \Cref{appendix:model_config}.

For both approaches, reconstruction and forecasting, we observe that the best window size doesn't correlate with the average anomaly length. 
This is surprising, as one would expect that anomalies covering entire time windows might be harder to detect, since the model might learn to reconstruct the anomalous window.
However, this might be a simplistic assumption, especially given the collective nature of these longer anomalies.

For \texttt{iTransformer-reco}, a smaller internal model size is preferred, as the largest internal model size did not produce remarkable results on any of the datasets.
This suggests that smaller model sizes successfully force compression of the input data, thus allowing the model to capture the average behaviour of the time series.
Consequently, this prevents the model from learning to reconstruct unusual patterns which might hint at anomalies.\\
Further, the step size is varied for the reconstruction strategy, where we observe a clear preference towards smaller step sizes.
A small step size creates more input windows for training, thus allowing the model to train on additional data while scanning the time series in more detail.
Therefore, the choice of the step size directly impacts the robustness of the learning model, as also observed in our parameter search.
However, we observed that comparable loss values and robustness can be achieved with larger step sizes, at the cost of longer training.


\subsection{Studies on Anomaly Label Extraction}
Next, we determine a preferred anomaly label extraction method for each dataset and parameter configuration.
For \texttt{iTransformer-fc}, all configurations achieve the best results with the same anomaly label method for a given dataset.
For \texttt{iTransformer-reco}, at least 80\% of the configurations on most datasets agreed on the same anomaly label method, which was the one that gave the best anomaly detection results.
The only exceptions to this were \texttt{SMAP} and \texttt{SWaT}.
For \texttt{SMAP}, all three anomaly label methods gave very similar results for half of the configurations; on the other half, the \textit{global} method dominated.
For \texttt{SWaT}, the \textit{local (majority voting)} and \textit{global} methods achieved very similar MCC scores.
\Cref{tab:anomalylabelmethod} details which anomaly label extraction method worked best for each multivariate dataset. 
It is interesting to note that the anomaly label method barely depends on the configuration of the model, but only on the dataset.

\begin{table}[th!]
    \caption{Anomaly label extraction giving the best results for reconstruction- or forecasting-based iTransformer models on each dataset. 
    The univariate datasets \texttt{SWaT\_1D} and \texttt{UCR} are omitted because all anomaly label extraction methods yield the same results.}
    \label{tab:anomalylabelmethod}
    \centering
    \begin{tabular}{@{}l|ll@{}}
    \toprule
    Dataset & \begin{tabular}[c]{@{}l@{}}Anomaly label extraction \\ for iTransformer-reco\end{tabular} & \begin{tabular}[c]{@{}l@{}}Anomaly label extraction \\ for iTransformer-fc\end{tabular} \\ \midrule
    Credit Card & local (incl. OR) & local (incl. OR) \\
    GECCO & local (incl. OR) & local (incl. OR) \\
    IEEECIS & local (incl. OR) & local (incl. OR) \\
    MSL & local (maj. voting) & global \\
    SMAP & local (maj. voting) & global \\
    SMD & local (incl. OR) & local (incl. OR) \\
    SWaT & local (maj. voting) & global \\
    WADI & global & global \\
    \bottomrule
    \end{tabular}%
\end{table}

We observe a clear pattern that datasets with shorter anomalous sequences (average anomaly length $< 100$) work best with the \textit{local (inclusive OR)}, whereas datasets with longer anomalous sequences prefer \textit{local (majority voting)} or \textit{global}.
Surprisingly, also \texttt{SMD}, which contains longer anomalous sequences, achieves the best MCC scores with \textit{local (inclusive OR)}.
Further studying the \texttt{SMD} test data, we find 35 anomalous sequences where ten of them are long, anomalous sequences ($\geq 100$ consecutive, anomalous time stamps).
However, there are also very short, almost point-like, anomalies in the test data which explains why the average anomaly length is relatively low.

Comparing the three anomaly label extraction methods, one notices that the point anomalies are only detected with the \textit{local (inclusive OR)} method, which seems to be the most sensitive to this anomaly type. 
This can be explained by the fact that point anomalies are very short and may only affect a single variate, whereas collective anomalies, by definition, involve several time stamps or dimensions.
Therefore, using the \textit{inclusive OR} method ensures that subtle anomalies, present in only one variate, are not overlooked.

Finally, we also observe that for larger anomalous sequences, the reconstruction-based \texttt{iTransformer} works better with \textit{local (majority voting)}, whereas the forecasting-based \texttt{iTransformer} prefers the \textit{global} approach.


\subsection{Influences of Anomalies in Training Data}
Given that the underlying assumption for both forecasting- and reconstruction-based TSAD relies on the rarity of anomalies in training data, it is crucial to investigate, how the presence of anomalies during training actually impacts the anomaly detection performance.
This is important as it is not possible to always guarantee the absence of anomalies in real-world applications.
Concretely, we study the effect of anomalies in training data on the example of the reconstruction-based \texttt{iTransformer} because it is the most flexible and efficient model of our selection.
First, we use the \texttt{Credit Card} and \texttt{GECCO} datasets, where anomaly labels are also available for the training data, thus allowing an explicit comparison of the anomaly detection performance when training with and without anomalies.
Second, further change the loss function during training, to check whether the loss functions presented in \Cref{sect:loss_fct} might mitigate this issue, as they may be less sensitive to outliers than the MSE.
However, for evaluation, we always employ the MSE as an anomaly score, as a high sensitivity to outliers is preferred in this case.

\begin{table}[th!]
\caption{MCC scores (averaged over five random seeds) for different loss functions during training, with and without anomalies in training data.}
\label{tab:loss_fct_wanomalies}
\centering
\begin{tabular}{@{}lc|ll@{}}
\toprule
Dataset  & \begin{tabular}[c]{@{}l@{}}Anomaly rate \\in train data\end{tabular} & Loss function & Best MCC \\ \midrule
Credit Card & 0.00\% & MSE & \textbf{0.241 $\pm$ 0.019} \\
 & & Huber & 0.220 $\pm$ 0.012 \\
 & & Soft-DTW & 0.230 $\pm$ 0.021 \\ \midrule
Credit Card & 0.19\% & MSE & 0.226 $\pm$ 0.020 \\
 &  & Huber & \textbf{0.234 $\pm$ 0.015} \\
 &  & Soft-DTW & 0.209 $\pm$ 0.012 \\ \midrule
GECCO & 0.00\% & MSE & 0.760 $\pm$ 0.004 \\
 & & Huber & 0.767 $\pm$ 0.010 \\
 & & Soft-DTW & \textbf{0.774 $\pm$ 0.018} \\ \midrule
GECCO & 1.43\% & MSE & 0.631 $\pm$ 0.039 \\
 & & Huber & 0.597 $\pm$ 0.063 \\
 & & Soft-DTW & \textbf{0.696 $\pm$ 0.024} \\
 \bottomrule
\end{tabular}%
\end{table}

\Cref{tab:loss_fct_wanomalies} presents the anomaly rates present in the training data and the MCC scores achieved by the \texttt{iTransformer-reco} when trained on these data.

For both datasets, we observe that the presence of anomalies in training data degrades the anomaly detection performance of the model.
This outcome is expected, as the model may learn to reconstruct anomalies, thereby invalidating the assumption that anomalies have a higher reconstruction error.
It is however remarkable that already such a low amount of anomalies suffices to affect the training process and deteriorate the reconstructed output.

Furthermore, for both datasets, employing a loss function other than the MSE seems to be beneficial when the training set includes anomalies. 
Therefore, in cases where it cannot be avoided to have anomalous instances in training data, alternative loss functions should be explored.
However, the lack of labelled training data in different datasets does not allow for a comprehensive comparison of alternative loss functions and their potential advantages when dealing with anomalies during training.
This further emphasises the need for more and better labelled benchmark datasets for TSAD.

For completeness, we also tested the effect of using alternative loss functions during training on the remaining datasets, where we observed a consistent anomaly detection performance across all loss functions.
The results can be found in \Cref{appendix:loss_fct}.

\subsection{Benchmarking Results for Transformer-based TSAD}
Next, we report the results of multiple transformer-based anomaly detection models for time series.
We investigated early stopping strategies based on the model's performance on a validation set, but noticed that the anomaly detection performance often deteriorated with longer training times.
Consequently, and also for computational efficiency, we chose to train all models for only ten epochs, repeating the procedure for five random seeds to obtain an estimate of a model's robustness.
We compare the baseline from \Cref{sect:baseline} and the \texttt{USAD} model from \Cref{sect:usad} against the two \texttt{iTransformer} implementations from \Cref{sect:iTransformer}, a reconstruction-based (vanilla) transformer and the \texttt{TranAD} model. 

For \texttt{TranAD}, we use the optimal configuration established 
in the corresponding publication (window size $W=10$, step size $S=1$, internal model size $D = 2\cdot N$ where $N$ is the dimension of the time series for \texttt{TranAD})~\cite{tuli_tranad_2022} and we proceed similarly for \texttt{USAD} (window size $W=10$, step size $S=5$, latent dimension = 5)~\cite{audibert_usad_2020}.

For the \texttt{vanilla Transformer}, we use the optimal parameter configuration for the reconstruction-based \texttt{iTransformer} model that was studied in \Cref{sect:iTransformer}, and apply the same configuration for a reconstruction-based transformer model.
The only remaining difference between the two models is the embedding, which is either \textit{normal} or \textit{inverted}.

Concerning the datasets, we use the same data for all models except for \texttt{SWaT}, where it was necessary to reduce the dataset size to a tenth of the original size, because all models except for \texttt{iTransformer-reco} struggled to process the full-sized one.
This is expected as the inverted embedding of the \texttt{iTransformer} was conceived to mitigate such issues, and reconstruction-based models often allow for more flexibility and efficiency in their implementation.
For \texttt{TranAD}, it was further necessary to restrict the variates of the input to the first 30 variates for all datasets due to computational limitations.

The MCC scores over five random seed repetitions on each dataset are shown in \Cref{tab:comparison_results}.

The baseline algorithm provides interesting insights into which datasets are almost `trivial' for anomaly detection, such as \texttt{SMD}, \texttt{SWaT\_1D} and maybe \texttt{MSL}, though to a lesser extent than the former two.
However, for the remaining datasets, the baseline does not offer any more information.

The \texttt{iTransformer} models dominate on eight out of ten datasets, the exceptions being the \texttt{SMAP} and \texttt{SWaT\_1D} datasets.
However, on both of these, \texttt{iTransformer-reco} reaches a score within the standard deviation of the highest score (achieved by \texttt{USAD}).
Furthermore, on \texttt{SWaT\_1D}, even the baseline algorithm achieves a high MCC score, suggesting that complex transformer models offer little to no advantage here. \\
The \texttt{iTransformer-reco} is the model achieving the best scores across the most datasets, but also demonstrates superior flexibility and efficiency, compared to the other models evaluated in this study.
This suggests that by inverting the embedding, one can achieve performance gains, highlighting the effectiveness of this simple yet impactful modification. \\
Interestingly, the \texttt{iTransformer-fc} is the best model on the \texttt{Credit Card} data, implying that predictive models might offer some advantages in identifying point anomalies. 

It is also noteworthy that on the univariate time series datasets, \texttt{SWaT\_1D} and \texttt{UCR}, the reconstruction-based \texttt{iTransformer} and \texttt{vanilla Transformer} achieve nearly identical MCC scores. 
This is expected as the multivariate attention mechanism of the \texttt{iTransformer} cannot be exploited on one-dimensional data and thus doesn't offer any advantage in learning feature correlations.

Finally, the \texttt{TranAD} and the \texttt{USAD} model struggle very much on multiple datasets.
On \texttt{Credit Card} and \texttt{IEEECIS}, both models fail at reconstructing the input and just return a constant value.
This also happens when applying \texttt{USAD} to \texttt{GECCO}, whereas \texttt{TranAD} fails to reconstruct \texttt{SWaT}.
Upon closer inspection of the \texttt{GECCO} and \texttt{IEEECIS} datasets, it seems that there are some trend changes in both of them, which hinder both models from correctly reconstructing the time series.
This leads to artificially high reconstruction errors in those regions of the time series where such trend changes are very noticeable and therefore the POT method converts these high reconstruction errors into (falsely) positive anomaly labels.

\begin{table}[ht]
\caption{MCC scores (averaged over five random seeds) for forecasting-based, reconstruction-based iTransformer, vanilla Transformer, TranAD and USAD model on various datasets. The baseline is deterministic, so only a single MCC score is reported. For the univariate time series \texttt{SWaT\_1D} and \texttt{UCR}, all anomaly label extraction methods yield the same results.}
\label{tab:comparison_results}
\centering
\begin{minipage}[t]{0.49\linewidth}
\centering
\resizebox{\linewidth}{!}{%
\begin{tabular}{@{}ll|ll@{}}
\toprule
Dataset & Model & Best MCC & \begin{tabular}[c]{@{}l@{}}Anomaly label\\ method\end{tabular} \\ \midrule
Credit Card & Baseline & 0.000 & local (maj. voting) \\
 & USAD &  0.000 $\pm$ 0.000 & local (maj. voting) \\ 
 & TranAD & 0.000 $\pm$ 0.000 & local (incl. voting) \\ 
 & Transformer & 0.254 $\pm$ 0.007 & local (incl. OR) \\
 & iTransformer-fc & \textbf{0.381 $\pm$ 0.001} & local (incl. OR) \\
 & iTransformer-reco & 0.241 $\pm$ 0.019 & local (incl. OR) \\
 \midrule
GECCO & Baseline & 0.027 & global \\
 & USAD &  0.000 $\pm$ 0.000 & local (maj. voting) \\ 
 & TranAD & 0.007 $\pm$ 0.013 & global \\ 
 & Transformer & 0.752 $\pm$ 0.01 & local (incl. OR) \\
 & iTransformer-fc & \textbf{0.806 $\pm$ 0.029} & local (incl. OR) \\
 & iTransformer-reco & 0.760 $\pm$ 0.004 & local (incl. OR) \\
 \midrule
IEEECIS & Baseline & 0.000 & local (maj. voting) \\
 & USAD &  0.000 $\pm$ 0.000 & local (maj. voting) \\ 
 & TranAD & 0.000 $\pm$ 0.000 & local (maj. voting) \\ 
 & Transformer & 0.612 $\pm$ 0.04 & local (incl. OR) \\
 & iTransformer-fc & 0.456 $\pm$ 0.008 & local (incl. OR) \\
 & iTransformer-reco & \textbf{0.629 $\pm$ 0.032} & local (incl. OR) \\
 \midrule
MSL & Baseline & 0.430 & local (incl. OR) \\
 & USAD & 0.626 $\pm$ 0.034 & global \\
 & TranAD & 0.154 $\pm$ 0.016 & local (incl. OR) \\ 
 & Transformer & 0.886 $\pm$ 0.016 & local (maj. voting) \\
 & iTransformer-fc & 0.866 $\pm$ 0.002 & global \\
 & iTransformer-reco & \textbf{0.935 $\pm$ 0.011} & local (maj. voting) \\
 \midrule
SMAP & Baseline & 0.000 & local (maj. voting) \\
 & USAD & \textbf{0.836 $\pm$ 0.014} & global \\
 & TranAD & 0.481 $\pm$ 0.16 & local (incl. OR) \\
 & Transformer & 0.557 $\pm$ 0.023 & local (incl. OR) \\
 & iTransformer-fc & 0.671 $\pm$ 0.002 & global \\
 & iTransformer-reco & 0.826 $\pm$ 0.091 & local (maj. voting) \\
\bottomrule
\end{tabular}%
}
\end{minipage}%
\hfill
\begin{minipage}[t]{0.48\textwidth}
\centering
\resizebox{\textwidth}{!}{%
\begin{tabular}{@{}ll|ll@{}}
\toprule
Dataset & Model & Best MCC & \begin{tabular}[c]{@{}l@{}}Anomaly label\\ method\end{tabular} \\ \midrule
SMD & Baseline & 0.924 & local (incl. OR) \\
 & USAD & 0.326 $\pm$ 0.056 & local (incl. OR) \\
 & TranAD & 0.726 $\pm$ 0.039 & local (incl. OR) \\
 & Transformer & 0.953 $\pm$ 0.005 & local (incl. OR) \\
 & iTransformer-fc & 0.913 $\pm$ 0.000 & local (incl. OR) \\
 & iTransformer-reco & \textbf{0.967 $\pm$ 0.010} & local (incl. OR) \\
 \midrule
SWaT & Baseline & 0.000 & all equal \\
 & USAD & 0.838 $\pm$ 0.023 & global \\
 & TranAD & 0.000 $\pm$ 0.000 & all equal \\ 
 & Transformer & 0.839 $\pm$ 0.011 & global \\
 & iTransformer-fc & 0.824 $\pm$ 0.052 & global \\ 
 & iTransformer-reco & \textbf{0.964 $\pm$ 0.004} & local (maj. voting) \\
 \midrule
SWaT\_1D & Baseline & 0.804 & all equal \\
 & USAD & \textbf{0.807 $\pm$ 0.002} & all equal \\
 & TranAD & 0.805 $\pm$ 0.006 & all equal \\ 
 & Transformer & 0.806 $\pm$ 0.002 & all equal \\
 & iTransformer-fc & 0.737 $\pm$ 0.050 & all equal \\
 & iTransformer-reco & 0.805 $\pm$ 0.006 & all equal \\
 \midrule
UCR & Baseline & 0.000 & all equal \\
 & USAD & 0.105 $\pm$ 0.234 & all equal \\
 & TranAD & 0.540 $\pm$ 0.306 & all equal \\ 
 & Transformer & 0.757 $\pm$ 0.019 & all equal \\
 & iTransformer-fc & 0.305 $\pm$ 0.274 & all equal \\
 & iTransformer-reco & \textbf{0.775 $\pm$ 0.010} & all equal \\
 \midrule
WADI & Baseline & 0.153 & local (incl. OR) \\
 & USAD & 0.411 $\pm$ 0.080 & global \\
 & TranAD & 0.530 $\pm$ 0.176 & global \\
 & Transformer & 0.804 $\pm$ 0.037 & global \\
 & iTransformer-fc & 0.676 $\pm$ 0.056 & global \\
 & iTransformer-reco & \textbf{0.825 $\pm$ 0.024} & global \\
\bottomrule
\end{tabular}%
}
\end{minipage}
\end{table}

\section{Summary and Outlook}
We investigated each stage of the end-to-end TSAD algorithm, including the generation of anomaly scores from anomaly detection models, the process of combining and converting these scores into anomaly labels, and the evaluation of the resulting labels using appropriate metrics. 

We explored the \texttt{iTransformer} ~\cite{liu_itransformer_2024} as a robust model for anomaly detection based on either reconstruction or forecasting, and analysed the impact of key parameters.
Especially, the reconstruction-based \texttt{iTransformer} proved to be flexible and efficient on datasets with various dimensions and anomaly types.
Additionally, we investigated methods for deriving anomaly labels from multi-dimensional anomaly scores generated by detection models and derived recommendations on how to find the best method given a dataset.

We further studied the impact of anomalous data in training on the anomaly detection performance and observed a clear deterioration even with very low anomaly rates.
It was possible to slightly mitigate this effect by employing loss functions different from the MSE.
However, due to the lack of labelled training data, it was only possible to conduct these studies on two datasets.
For a more exhaustive evaluation of the effects of anomalies during training, more such datasets are needed, specially with higher diversity of anomalies, or ways to control the rate of anomalies present in training data.
 
Finally, we compared various transformer-based models for TSAD across a diverse set of datasets, ensuring a broad representation of applications. 
We concluded that the \texttt{iTransformer} model demonstrates a strong suitability for TSAD, effectively distinguishing anomalous patterns from normal temporal behaviour.
This highlights the performance impact of the inverted embedding and raises questions about the necessity of applying more complex architectural modifications to transformers.

\begin{acksection}
We acknowledge funding from the European Union Horizon 2020 research and innovation programme, call H2020-MSCA-ITN-2020, under Grant Agreement n. 956086.  
This work benefitted from the source code provided by~\citet{tuli_tranad_2022} and ~\citet{liu_itransformer_2024}.

We would like to thank Shubham Gupta for the insightful discussions, and continuous support throughout the development of this work. 
We also gratefully acknowledge Pierre Feillet for the helpful discussions and perspectives shared during the course of this project.
\end{acksection}

\clearpage

\appendix

\section{Descriptions of Datasets for TSAD}
\label{appendix:data_more}
The following list offers some more details on each dataset used in this paper:

\begin{enumerate}
    \item \texttt{Credit Card} data: The dataset describes credit card transactions by European cardholders in September 2013. It has been prepared by a research collaboration of Worldline and the Machine Learning Group of Université Libre de Bruxelles and is now accessible in the form of a kaggle challenge~\cite{pozzolo_calibrating_2015}. For confidentiality reasons, the initial features have been anonymised by applying a PCA transformation.
    \item \texttt{GECCO} data: This set contains real-world data collected for \textit{'GECCO 2018 Industrial Challenge'}~\cite{bartz-beielstein_internet_2018}.  
    The time series has nine variates which describe the temperature, the Chlorine dioxide concentration, the pH value of the water etc.
    \item \texttt{IEEECIS} data: The dataset consists of e-commerce data provided by Vesta Corporation for the IEEE-CIS fraud detection kaggle challenge~\cite{howard_ieee-cis_2019}. The fraud-dataset-benchmark repository ~\cite{grover_fraud_2023} provides a cleaned-up version of this dataset with a reduced number of features and including a user ID. From there, we create multivariate user time series tracking the transactions of one individual user over time.
    \item Mars Science Laboratory (\texttt{MSL}) data: The set consists of real-world data collected by sensors of NASA's Mars Rover ~\cite{hundman_detecting_2018}. 
    We only kept the sequences (A-4, C-2, T-1) that were non-trivial, as mentioned in ~\cite{nakamura_merlin_2020}.
    \item Soil Moisture Active Passive (\texttt{SMAP}) data: A real-world dataset that has been collected by the NASA satellite SMAP~\cite{hundman_detecting_2018}.
    \item Server Machine Dataset (\texttt{SMD}): The dataset contains real-world data from an internet company, which has been collected over five weeks and incident reports have been labelled by experts~\cite{su_robust_2019}. It contains information like the resources used by individual machines in a cluster. Based on comments from ~\cite{wu_current_2021, tuli_tranad_2022}, we only include four individual time series with less trivial anomalies (namely sequences `machine-1-1', `machine-2-1', `machine-3-2', `machine-3-7').
    \item Secure Water Treatment (\texttt{SWaT}) data: It contains real data collected by a water treatment plant over 11 days of continuous operation: seven under normal operation and four days with attack scenarios~\cite{mathur_swat_2016}. We use the updated version, which excludes the first 30 minutes during which the water tank is filled.
    Since some algorithms struggle with the fully-sized dataset, we use a second version of the set that has been down sampled to one tenth of its initial volume.
    \item \texttt{SWaT\_1D} data: A univariate time series based on \texttt{SWaT} data which has been prepared and provided for an anomaly detection tutorial~\cite{julienau_anomaly_2020}.
    \item Hexagon ML/UCR Time Series Anomaly Archive (\texttt{UCR}) data: The dataset contains multiple one-dimensional time series that are based on real data. We focus on time series where anomalies from natural sources (the InternalBleeding and ECG datasets) are artificially introduced in those time series~\cite{wu_current_2021}. 
    \item Water Distribution (\texttt{WADI}) data: Similarly to \texttt{SWaT}, \texttt{WADI} contains data from 123 sensors and actuators of a real water plant with 14 days of normal operation and two days with various attack scenarios~\cite{ahmed_wadi_2017}. For efficiency reasons, the dataset has been downsampled to one-tenth of its initial volume.
\end{enumerate}

\clearpage
\section{iTransformer Model Configurations}
\label{appendix:model_config}

To determine the best configuration for the \texttt{iTransformer} model in the context of reconstruction-based and forecasting-based anomaly detection, we relate the window size $W$ to the step size $S$ and to the internal model size $M$.
For all other parameters, such as batch size (fixed at $32$), learning rate (fixed at $10^{-4}$) and training epochs (fixed at $10$), we stick to the default values given in the \texttt{iTransformer} publication~\cite{liu_itransformer_2024}.
For the other parameters, we use this parameter search as an opportunity to investigate the relations between a model's internal size and the window and step size that determine how the model reads the data.

We assume that for a specific dataset, the users have some domain knowledge and can estimate the average anomaly length $a$ and the minimal anomaly length $b$. 
This estimation can be based on a small, labelled test set, as that is needed for the evaluation of the anomaly detection method any ways.
For our case, we use the values provided in \Cref{tab:public_data}. \\
Next, for every dataset, we determine a big and a small window size $W_+$ and $W_-$, respectively.
First, $W_+$ is chosen as the multiple of 50 that is greater than the average anomaly length of the dataset.
Second, $W_-$ is set to the maximum value between $[a/2]$, where the square brackets denote rounding half up to the nearest integer, and $10$. The lower bound of $10$ is imposed because smaller time windows start to undermine the sequential nature of the data.
Finally, we also use $W_{default}=96$, which is based on the \texttt{iTransformer} paper.
For reconstruction-based anomaly detection, there are again two options $S_+=[W/2]$ and $S_-=[W/10]$ for the step size.
For forecasting-based anomaly detection, we set $S=1$ as typically done for forecasting-based models~\cite{zamanzadeh_darban_deep_2024}. 
We further fix $M=2$ for efficiency reasons.\\
Last, the model internal size $M$ (denoted as \textit{d\_model} in the \texttt{iTransformer} paper) can also take two different values, either it is set equal to the window size, i.e. $M_+=W$, or to a fifth of the window size, i.e. $M_-=[W/5]$.

This approach leaves us with 12 possible \texttt{iTransformer} model configurations for reconstruction-based anomaly detection and 6 configurations for forecasting-based anomaly detection, and this for each dataset.
Each configuration is repeated with five different random seeds to gauge the model's robustness.
We also use this parameter search to compare the impact of the different anomaly label extraction strategies presented in \Cref{anomaly_label_methods}.

For each parameter configuration and each of the three anomaly labelling methods, the MCC score is computed, and the configuration with the highest MCC score is selected.
If the best MCC score over five repetitions plus or minus its standard deviation overlaps with the MCC score from other configurations, the configuration with the smallest model size is chosen.
Next, if there are still multiple optimal configurations remaining, the one with the smallest step size is taken.
This typically also coincides with the configuration with the smaller standard deviation, i.e. the more robust choice.

The final \texttt{iTransformer} configuration choices on each dataset are given below: \Cref{tab:config_reco} lists the parameter configurations for reconstruction-based anomaly detection and \Cref{tab:configs_fc} for forecasting-based anomaly detection.

\clearpage
\begin{table}[ht!]
\caption{Best iTransformer configurations for reconstruction-based anomaly detection with window size $W$, step size $S$ and internal model size $M$ for every dataset. For the univariate time series \texttt{SWaT\_1D} and \texttt{UCR}, all anomaly label extraction methods yield the same results.}
\label{tab:config_reco}
\centering
\begin{tabular}{@{}l|c|cccl@{}}
\toprule
Dataset & Best MCC score & $W$ & $S$ & $M$ & \begin{tabular}[c]{@{}l@{}}Anomaly label\\ method\end{tabular} \\ \midrule
Credit Card & 0.241 $\pm$ 0.019 & 10 & 1 & 2 & local (incl. OR) \\
GECCO & 0.760 $\pm$ 0.004 & 12 & 1 & 2 & local (incl. OR) \\
IEEECIS & 0.629 $\pm$ 0.032 & 10 & 1 & 10 & local (incl. OR) \\
MSL & 0.935 $\pm$ 0.011 & 96 & 10 & 96 & local (maj. voting) \\
SMAP & 0.826 $\pm$ 0.091 & 15 & 2 & 15 & local (maj. voting) \\
SMD & 0.967 $\pm$ 0.010 & 350 & 35 & 70 & local (incl. OR) \\
SWaT & 0.964 $\pm$ 0.004 & 51 & 5 & 10 & local (maj. voting) \\
SWaT\_1D & 0.805 $\pm$ 0.006 & 10 & 5 & 2 & - \\
UCR & 0.775 $\pm$ 0.010 & 10 & 5 & 2 & - \\
WADI & 0.825 $\pm$ 0.024 & 44 & 4 & 44 & global \\
\bottomrule
\end{tabular}
\end{table}

\begin{table}[ht!]
\caption{Best iTransformer configurations for forecasting-based anomaly detection with window size $W$ for every dataset, while the step size $S=1$ and internal model size $M=2$ are fixed. For the univariate time series \texttt{SWaT\_1D} and \texttt{UCR}, all anomaly label extraction methods yield the same results.}
\label{tab:configs_fc}
\centering
\begin{tabular}{@{}l|c|cl@{}}
\toprule
Dataset & Best MCC score & $W$ & \begin{tabular}[c]{@{}l@{}}Anomaly label method\end{tabular} \\ \midrule
Credit Card & 0.381 $\pm$ 0.001 & 96 & local (incl. OR) \\
GECCO & 0.806 $\pm$ 0.029 & 12 & local (incl. OR) \\
IEEECIS & 0.456 $\pm$ 0.008 & 10 & local (incl. OR) \\
MSL & 0.866 $\pm$ 0.001 & 10 & global \\
SMAP & 0.671 $\pm$ 0.002 & 15 & global \\
SMD & 0.913 $\pm$ 0.000 & 10 & local (incl. OR) \\
SWaT & 0.824 $\pm$ 0.052 & 51 & global \\
SWaT\_1D & 0.737 $\pm$ 0.050 & 50 & - \\
UCR & 0.305 $\pm$ 0.274 & 96 & - \\
WADI & 0.676 $\pm$ 0.056 & 44 & global \\ 
\bottomrule
\end{tabular}%
\end{table}

\clearpage
\section{Results with Alternative Loss Functions}
\label{appendix:loss_fct}
We study the influence of changing the loss function on the example of the reconstruction-based \texttt{iTransformer}.
Namely, we compare the anomaly detection performance using three different training losses: MSE, Huber, and Soft-DTW.
Regardless of the training loss, during testing, the anomaly score is always computed with the MSE.
This choice is motivated by two reasons.
First, models are trained with the Huber loss and evaluated with the MSE, because the MSE is more sensitive to outliers, which should be avoided during training but not during testing.
Second, the Soft-DTW loss does not produce an element-wise loss that could serve as an anomaly score.
For each dataset, the \texttt{iTransformer} model uses the optimal parameter configuration determined previously and is trained over ten epochs.

The results when training with each of the three loss functions are presented in \Cref{tab:loss_fct}.

\begin{table}[ht!]
\centering
\caption{MCC scores (averaged over five random seeds) for reconstruction-based iTransformer using either the mean-squared error (MSE), Huber loss or the soft dynamic time warping (Soft-DTW) loss during training and the MSE during testing. For the univariate time series \texttt{SWaT\_1D} and \texttt{UCR}, all anomaly label extraction methods yield the same results.}
\label{tab:loss_fct}
\begin{tabular}{@{}l|l|ll@{}}
\toprule
Dataset & Loss function & Best MCC & \begin{tabular}[c]{@{}l@{}}Anomaly label\\ method\end{tabular} \\ \midrule
Credit Card & MSE & \textbf{0.241 $\pm$ 0.019} & local (incl. OR) \\
 & Huber & 0.220 $\pm$ 0.012 & local (incl. OR) \\
 & Soft-DTW & 0.230 $\pm$ 0.021 & local (incl. OR) \\ \midrule
GECCO & MSE & 0.760 $\pm$ 0.004 & local (incl. OR) \\
 & Huber & 0.767 $\pm$ 0.010 & local (incl. OR) \\
 & Soft-DTW & \textbf{0.774 $\pm$ 0.018} & local (incl. OR) \\ \midrule
IEEECIS & MSE &\textbf{0.629 $\pm$ 0.032} & local (incl. OR) \\
 & Huber & 0.532 $\pm$ 0.129 & local (incl. OR) \\
 & Soft-DTW & 0.567 $\pm$ 0.071 & local (incl. OR) \\ \midrule
MSL & MSE & 0.935 $\pm$ 0.011 & local (maj. voting) \\
 & Huber &\textbf{0.938 $\pm$ 0.006 }& local (maj. voting) \\
 & Soft-DTW & 0.933 $\pm$ 0.009 & local (maj. voting) \\ \midrule
SMAP & MSE &\textbf{0.826 $\pm$ 0.091} & local (maj. voting) \\
 & Huber & 0.777 $\pm$ 0.066 & local (maj. voting) \\
 & Soft-DTW & 0.714 $\pm$ 0.073 & local (incl. OR) \\ \midrule
SMD & MSE &\textbf{0.967 $\pm$ 0.010} & local (incl. OR) \\
 & Huber & 0.956 $\pm$ 0.019 & local (incl. OR) \\
 & Soft-DTW & 0.955 $\pm$ 0.020 & local (incl. OR) \\ \midrule
SWaT & MSE & 0.964 $\pm$ 0.004 & local (maj. voting) \\
 & Huber & \textbf{0.967 $\pm$ 0.006} & local (maj. voting) \\
 & Soft-DTW & 0.963 $\pm$ 0.020 & local (maj. voting) \\ \midrule
SWaT\_1D & MSE &\textbf{0.810 $\pm$ 0.000} & - \\
 & Huber & 0.806 $\pm$ 0.002 & - \\
 & Soft-DTW & 0.806 $\pm$ 0.005 & - \\ \midrule
UCR & MSE & \textbf{0.775 $\pm$ 0.010} & - \\
 & Huber & 0.716 $\pm$ 0.115 & - \\
 & Soft-DTW & 0.758 $\pm$ 0.041 & - \\ \midrule
WADI & MSE & 0.825 $\pm$ 0.024 & global \\
 & Huber &\textbf{0.847 $\pm$ 0.004} & global \\
 & Soft-DTW & 0.774 $\pm$ 0.051 & global \\ \bottomrule 
\end{tabular}%
\end{table}

\clearpage

\bibliographystyle{unsrtnat}
\bibliography{references}

\end{document}